\documentclass[letterpaper, 10 pt, conference]{ieeeconf}  

\IEEEoverridecommandlockouts                              

\usepackage{graphics} 
\usepackage{epsfig} 
\usepackage{mathptmx} 
\usepackage{times} 
\usepackage{amsmath} 
\usepackage{amssymb}  
\usepackage[T1]{fontenc}
\usepackage{cite}
\usepackage{pifont}

\title{\LARGE \bf
GAM-Depth: Self-Supervised Indoor Depth Estimation Leveraging a Gradient-Aware Mask and Semantic Constraints
}

\author{Anqi Cheng$^{1}$, Zhiyuan Yang$^{1}$, Haiyue Zhu$^{2}$ and Kezhi Mao$^{1}$$^{*}$
\\
\thanks{$^{*}$Corresponding author}
\thanks{$^{1}$A.  Cheng, Z. Yang and K. Mao are with the School of Electrical and Electronic Engineering, Nanyang Technological University, Singapore (e-mail: \{anqi002, yang0674, ekzmao\}@ntu.edu.sg).}
\thanks{$^{2}$H.~Zhu is with Singapore Institute of Manufacturing Technology (SIMTech), Agency for Science, Technology and Research (A*STAR), 2 Fusionopolis Way, Singapore, 138634, Singapore.(e-mail: zhu\_haiyue@simtech.a-star.edu.sg)}
}

\begin{document}

\maketitle
\thispagestyle{empty}
\pagestyle{empty}

\begin{abstract}

Self-supervised depth estimation has evolved into an image reconstruction task that minimizes a photometric loss. While recent methods have made strides in indoor depth estimation, they often produce inconsistent depth estimation in textureless areas and unsatisfactory depth discrepancies at object boundaries. To address these issues, in this work, we propose GAM-Depth, developed upon two novel components: gradient-aware mask and semantic constraints. The gradient-aware mask enables adaptive and robust supervision for both key areas and textureless regions by allocating weights based on gradient magnitudes. 
The incorporation of semantic constraints for indoor self-supervised depth estimation improves depth discrepancies at object boundaries, leveraging a co-optimization network and proxy semantic labels derived from a pretrained segmentation model. Experimental studies on three indoor datasets, including NYUv2, ScanNet, and InteriorNet, show that GAM-Depth outperforms existing methods and achieves state-of-the-art performance, signifying a meaningful step forward in indoor depth estimation. Our code will be available at \verb|https://github.com/AnqiCheng1234/GAM-Depth|.

\end{abstract}

\section{INTRODUCTION}

Depth estimation is a pivotal task in the field of robot vision, serving as a crucial component task for several applications, including robot navigation~\cite{pan2021mulls},~\cite{kareer2023vinl},~\cite{9812423}, robot manipulation~\cite{gu2017deep},~\cite{zhu2020grasping},~\cite{takahashi2021uncertainty}, and augmented reality~\cite{liu2018interactive}. Although supervised methods~\cite{eigen2014depth},~\cite{liu2015deep},~\cite{laina2016deeper},~\cite{fu2018deep},\cite{lee2019big} have made progress in depth estimation, they demand a large number of labeled RGB-depth pairs during model training, which may not always be viable. To address this limitation, self-supervised learning methods~\cite{garg2016unsupervised},~\cite{godard2017unsupervised},~\cite{zhou2017unsupervised},~\cite{godard2019digging},~\cite{almalioglu2019ganvo},~\cite{wang2020geometric} have emerged as popular solutions, as they only demand stereo pairs or monocular sequences. Most of these methods reframe depth estimation as an image reconstruction task, estimating the depth and camera poses simultaneously to perform inverse warping on RGB frames. 

\begin{figure}
\centering
\includegraphics[width=3.4in]{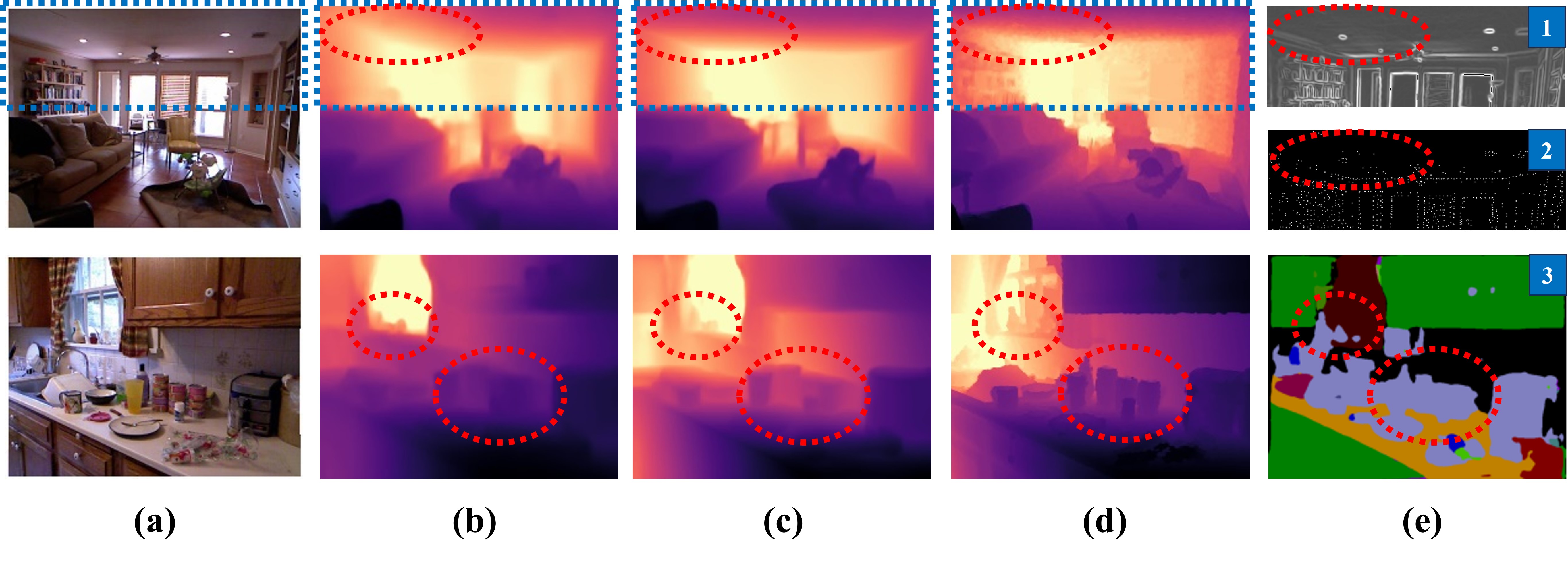}
\caption{Depth estimation comparisons. (a) Input, (b) StructDepth~\cite{li2021structdepth}, (c) Ours, (d) Ground truth, (e) 1. our proposed gradient-aware mask within the blue box region. The gray level of each pixel corresponds to its weight, with closer to white indicating higher weights. 2. the binary mask of keypoints~\cite{li2021structdepth} with textureless regions being neglected. 3. Semantic constraints with colors representing proxy semantic labels.}
\label{fig:1}
\end{figure}

Despite efficacy of the self-supervised methods in outdoor autonomous driving scenarios (e.g., KITTI\cite{geiger2012we}), their performance tends to decline in indoor environments (e.g., NYUv2\cite{silberman2012indoor}). This is partly attributed to the presence of \textit{large textureless regions} in indoor environments, such as walls, ceilings, and floors, whose primary supervision signal (photometric loss) becomes unreliable due to lack of texture information~\cite{yu2020p}. To address this issue, notable works such as P\textsuperscript{2}Net~\cite{yu2020p} and StructDepth~\cite{li2021structdepth} restrict the calculation of photometric loss to texture regions represented by keypoints, preventing non-texture areas from "polluting" the photometric loss. Although these attempts produce improved results, we argue that ignoring textureless regions during model training may not be an ideal solution. As illustrated in Fig~\ref{fig:1}(b), the absence of supervision in textureless regions can result in depth inconsistencies in these areas. Such inconsistent depth prediction for textureless regions such as floors and walls may hinder the robot's obstacle detecting and motion planning processes during indoor navigation~\cite{pan2021mulls},~\cite{kareer2023vinl},~\cite{9812423}.

Another issue of the current methods is the inaccurate prediction of \textit{depth discrepancies at object boundaries}, particularly with small and transparent objects. As illustrated in Fig~\ref{fig:1}(b), blurred boundaries are produced due to inaccurate prediction of depth discrepancies, which could potentially lead to complications in tasks such as robot-arm grasping and robotic manipulation~\cite{gu2017deep},~\cite{zhu2020grasping},~\cite{takahashi2021uncertainty}. We believe this problem is due to the limited capability of the feature extractor in the depth backbone, failing to capture object boundary information. To address this issue, we incorporate semantic constraints into feature extractor, aiming to improve the depth discrepancies at object boundaries.

In this work, we propose GAM-Depth, a comprehensive solution that attempts to provide adaptive and robust supervision signals for both textureless regions and key areas meanwhile incorporating semantic constraints into depth boundary predictions. Firstly, we develop a gradient-aware mask that allocates smooth weights to all pixels based on the magnitude of gradient, without having to identify keypoints by thresholding the gradient magnitude. This will balance the supervision strength in photometric loss for both the texture-rich and texture-less areas. 
Secondly, drawing inspiration from~\cite{tosi2020distilled},~\cite{seichter2021efficient}, we incorporate semantic constraints into the depth estimation backbone, aiming for clear depth boundaries. This is accomplished by sharing a feature extractor between a semantic segmentation task and our depth prediction task. 
Recognizing the crucial role of pose estimation in indoor depth estimation, our model synergizes residual pose optimization~\cite{ji2021monoindoor} to ensure that depth predictions are optimized in the correct direction. 

Experiments are conducted on three indoor datasets: NYUv2~\cite{silberman2012indoor}, ScanNet~\cite{dai2017scannet}, and InteriorNet~\cite{li2018interiornet}. The results demonstrate that GAM-Depth outperforms existing state-of-the-art methods and improves the $\delta_1$ threshold accuracy by a large margin. Fig.~\ref{fig:1}(b) and (c) show that our approach excels in predicting smooth depth in textureless regions and clear depth boundaries, highlighted in red boxes. The main contributions are as follows:
\begin{itemize}
\item We propose a gradient-aware mask in the photometric loss calculation, providing adaptive and robust supervision for both key areas and textureless regions. This gradient-aware mask can be combined with other self-supervised depth estimation methods to improve their performance. 
\item We incorporate semantic constraints into depth estimation through a co-optimization network and proxy semantic labels of a pretrained lightweight segmentation model. Our approach pioneers the use of semantic constraints in indoor self-supervised depth estimation.
\item The proposed model achieves state-of-the-art results on NYUv2~\cite{silberman2012indoor}, ScanNet~\cite{dai2017scannet}, and InteriorNet~\cite{li2018interiornet}.
\end{itemize}

\section{RELATED WORK}

\subsection{Self-supervised Monocular Depth Estimation}

Although existing supervised monocular depth estimation methods~\cite{eigen2014depth},~\cite{laina2016deeper},~\cite{fu2018deep},~\cite{lee2019big},~\cite{yuan2022neural},~\cite{patil2022p3depth} have made significant progress, they demand large quantity of annotated data, which may not be always available. To address this limitation of the supervised methods, ~\cite{garg2016unsupervised} introduced the first self-supervised depth estimation method that takes a stereo pair as the input. Godard et al.~\cite{godard2017unsupervised} introduced a left-right consistency loss, ensuring both images in the stereo pair had consistent depth accuracy. The paradigm was then further revolutionized by~\cite{zhou2017unsupervised}, which demonstrated that promising results could be achieved with just monocular video sequences. As one of the most representative self-supervised depth estimation methods, Monodepth2~\cite{godard2019digging} generates accurate and sharp depth maps leveraging an encoder-decoder network with skip connections. The main idea is to employ a minimum reprojection loss to handle occluded and out-of-view pixels while dealing with moving objects with automasking. Following this groundbreaking work, several attempts~\cite{guizilini20203d},~\cite{watson2021temporal} have been made to enhance depth estimation performance for autonomous driving scenarios. ~\cite{oquab2023dinov2} presented a novel approach that combines discriminative self-supervised learning techniques for robust feature learning on a large, curated dataset. 

Most aforementioned methods were developed in the context of outdoor scenes. For indoor environments, MovingIndoor~\cite{zhou2019moving} applied optical flow detection to provide an additional supervisory signal,~\cite{bian2021auto} introduced an auto-rectify network targeting the noise emerging from rotational variations between consecutive frames. P\textsuperscript{2}Net~\cite{yu2020p} first extracts keypoints by detecting pixels with large gradients, and then limit the photometric loss calculation exclusively to supported regions of the keypoints only. Expanding on this, StructDepth~\cite{li2021structdepth} explored indoor structural regularities by a Manhattan normal detection module and a planar region detection module. This ensured the predicted surface normals aligned with main directions while simultaneously generating planar segments. MonoIndoor~\cite{ji2021monoindoor} proposed depth estimation accuracy via a residual pose estimation module, recognizing the challenges of predicting accurate relative poses due to complex indoor rotational motions. Extending this idea further, HI-Net~\cite{wu2022hi} incorporated a homography estimation module and an iterative refinement process. SC-DepthV3~\cite{sun2022sc} and DistDepth~\cite{wu2022toward} made use of an off-the-shelf depth estimator to generate a single-image depth prior.

\subsection{Semantic Constraints for Depth Estimation}

Recent works in depth estimation have witnessed incorporation of semantic constraints for performance enhancement. Hambarde et al.~\cite{hambarde2020depth} proposed a semantic-to-depth generative adversarial network, while~\cite{zhu2020edge} minimized the border consistency between depth and segmentation. Wang et al.~\cite{wang2020sdc} leveraged both semantic and instance segmentation to predict category-agnostic and instance-level depth maps. Klinger et al.~\cite{klingner2020self} performed cross-task guidance by incorporating semantic segmentation in a supervised mode. A shared feature extractor is employed to close the gap between the two tasks. Our method is inspired by~\cite{tosi2020distilled}, which jointly predicted depth, relative pose, optical flow, semantic masks, and object labels to perform comprehensive scene understanding, utilizing a proxy semantic network. Unlike~\cite{tosi2020distilled}, which detects dynamic classes to exclude moving objects in autonomous driving scenes, we focus on enhancing depth estimation with semantic constraints in indoor environments.

\section{Method}

\subsection{Method Overview}

\begin{figure*}
\centering
\includegraphics[width=7in]{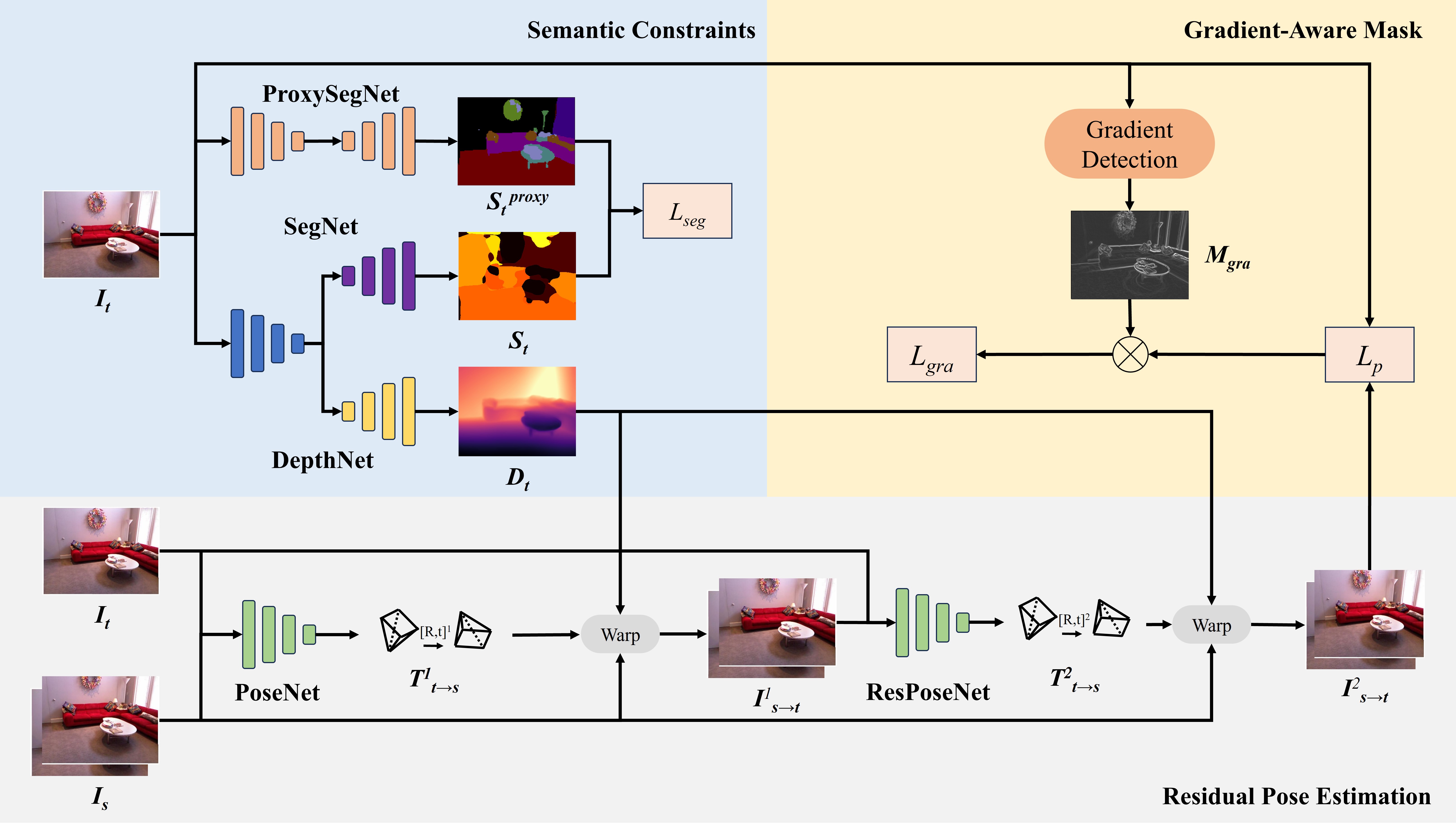}
\caption{Overall training pipeline of GAM-Depth. Given a target frame $\emph{\textbf{I}}_t$, DepthNet estimates its depth map $\emph{\textbf{D}}_t$ and SegNet predicts its semantic labels $\emph{\textbf{S}}_t$. DepthNet shares its encoder with SegNet. ProxySegNet predicts proxy semantic labels $\emph{\textbf{S}}_t^{proxy}$ to supervise the training of GAM-Depth through segmentation loss $L_{seg}$. The reference frame $\emph{\textbf{I}}_s$ is warped to $\emph{\textbf{I}}_t$'s view through PoseNet and ResPoseNet proposed by MonoIndoor~\cite{ji2021monoindoor}. Our gradient-aware mask $\emph{\textbf{M}}_{gra}$ is generated by a gradient detection method. Finally, a gradient-aware photometric loss $L_{gra}$ is calculated as the multiplication of the photometric loss $L_{p}$ between $\emph{\textbf{I}}_t$ and the warped frame $\emph{\textbf{I}}_{s \rightarrow t}$ and $\emph{\textbf{M}}_{gra}$.}
\label{fig:overall pipeline}
\end{figure*}
In this work, we develop GAM-Depth, comprising two new components: gradient-aware mask and semantic constraints. The gradient-aware mask $\emph{\textbf{M}}_{gra}$ enhances depth estimation consistency by providing robust supervision signals for both textureless regions and texture-rich areas (introduced in Sec.~\ref{sec:gradient-aware mask}) while the semantic constraints $L_{seg}$ improve boundary depth discrepancies by merging low-level details with high-level scene comprehension (introduced in Sec.~\ref{sec:semantic constraints}). For textureless regions with large proportions in indoor environments, we assign small and smooth weights instead of excluding them from supervision~\cite{yu2020p},~\cite{li2021structdepth}. For texture-rich regions with abrupt gradient changes, the gradient-aware mask is also able to assign smooth weights adaptively instead of a uniform allocation~\cite{yu2020p},~\cite{li2021structdepth}. For object boundaries, under the guidance of the semantic constraints, clearer depth discrepancies are produced.

The overall training pipeline of GAM-Depth is shown in Fig.~\ref{fig:overall pipeline}. In our self-supervised training architecture, monocular sequences serve as the exclusive inputs. DepthNet estimates the depth map $\emph{\textbf{D}}_t$ of the target frame $\emph{\textbf{I}}_t$, while the SegNet predicts its semantic labels $\emph{\textbf{S}}_t$ simultaneously. These two networks share an encoder for a comprehensive understanding of the scene. ProxySegNet, which is an off-the-shelf lightweight semantic segmentation model~\cite{nekrasov2018light}, provides proxy semantic labels $\emph{\textbf{S}}_t^{proxy}$ to guide the training of SegNet. PoseNet and ResPoseNet are utilized to generate accurate relative poses $\emph{\textbf{T}}_{\emph{t} \rightarrow \emph{s}}$ between the target frame $\emph{\textbf{I}}_t$ and the reference frame $\emph{\textbf{I}}_s$. With $\emph{\textbf{D}}_t$ and $\emph{\textbf{T}}_{\emph{t} \rightarrow \emph{s}}$ being predicted, following~\cite{zhou2017unsupervised}, the warping process 
can be described as:
\begin{equation}
   \emph{p}_{\emph{s}} \thicksim \emph{\textbf{K}} \emph{\textbf{T}}_{\emph{t} \rightarrow \emph{s}} \emph{\textbf{D}}_\emph{t}(p_t) \emph{\textbf{K}}^{-1} \emph{p}_\emph{t}
\end{equation}
where $p_t$ and $p_s$ are pixels of $\emph{\textbf{I}}_t$ and $\emph{\textbf{I}}_s$, and  $\emph{\textbf{K}}$ denotes the camera intrinsic matrix. The warped frame $\emph{\textbf{I}}_{s \rightarrow t}$ can be reconstructed following~\cite{zhou2017unsupervised}:
\begin{equation}
    \emph{\textbf{I}}_{s \rightarrow t} = \emph{\textbf{I}}_t \left\langle  proj  (\emph{\textbf{D}}_t, \emph{\textbf{T}}_{t \rightarrow s}, \emph{\textbf{K}}) \right\rangle
\end{equation}
where $\left\langle  \cdot \right\rangle$ is a locally sub-differentiable bilinear sampling operator~\cite{jaderberg2015spatial}. Our gradient-aware mask $\emph{\textbf{M}}_{gra}$ is generated through a smoothing function on the gradient magnitude,  which contributes to the photometric loss $L_{p}$ calculation.

\subsection{Gradient-Aware Mask}
\label{sec:gradient-aware mask}

In self-supervised training, the absence of ground truth depth information necessitates the reliance on the photometric loss $L_p$~\cite{godard2017unsupervised}, which is a weighted combination of the L1 and SSIM loss~\cite{wang2004image}:
\begin{equation}
   L_p(\emph{\textbf{I}}_t, \emph{\textbf{I}}_{s \rightarrow t}) = (1-\alpha)L_1 (\emph{\textbf{I}}_t, \emph{\textbf{I}}_{s \rightarrow t}) + \alpha SSIM(\emph{\textbf{I}}_t,\emph{\textbf{I}}_{s \rightarrow t})
\end{equation}
where $\emph{\textbf{I}}_t$, $\emph{\textbf{I}}_s$, and $\emph{\textbf{I}}_{s \rightarrow t}$ are the target frame, reference frame, and warped frame, respectively. $\alpha$ denotes the weight of SSIM loss~\cite{wang2004image}.

The minimum reprojection loss $L_{min}$, as proposed by Monodepth2~\cite{godard2019digging}, proves to be effective in self-supervised learning for its ability to mask out occluded and out-of-view pixels:
\begin{equation}
   L_{min} = \sum_{p_t} \min_s \left( L_p (\emph{\textbf{I}}_t, \emph{\textbf{I}}_{s \rightarrow t}) \right)
\end{equation}
where $s$ is the index of reference frames.

However, $L_{min}$ often fails in indoor environments with the presence of large textureless regions. To address this issue, P\textsuperscript{2}Net~\cite{yu2020p} and StructDepth~\cite{li2021structdepth} first extract keypoints~\cite{engel2017direct} based on pixel gradient magnitude thresholding, and then warp to calculate $L_{min}$ within the local support regions of these keypoints. Despite the success of these approaches, we identify two limitations: a lack of robustness in keypoints selection and an absence of supervision in textureless regions. To address these limitations, we propose a gradient-aware mask $\emph{\textbf{M}}_{gra}$ and a gradient-aware photometric loss $L_{gra}$:
\begin{equation}
    \emph{\textbf{M}}_{gra} = \beta + \frac{1-\beta} {1 + e^{-\gamma_1 \cdot \emph{\textbf{m}} + \gamma_2}}
\end{equation}
\begin{equation}
   L_{gra} = \sum_{p_t} \emph{\textbf{M}}_{gra}(p_t) \min_s \left( L_p (\emph{\textbf{I}}_t, \emph{\textbf{I}}_{s \rightarrow t}) \right)
\end{equation}
where $\emph{\textbf{m}}$ denotes pixel gradient magnitude, and $\beta$, $\gamma_1$, and $\gamma_2$ are hyperparameters.

\begin{figure}
\centering
\includegraphics[width=3in]{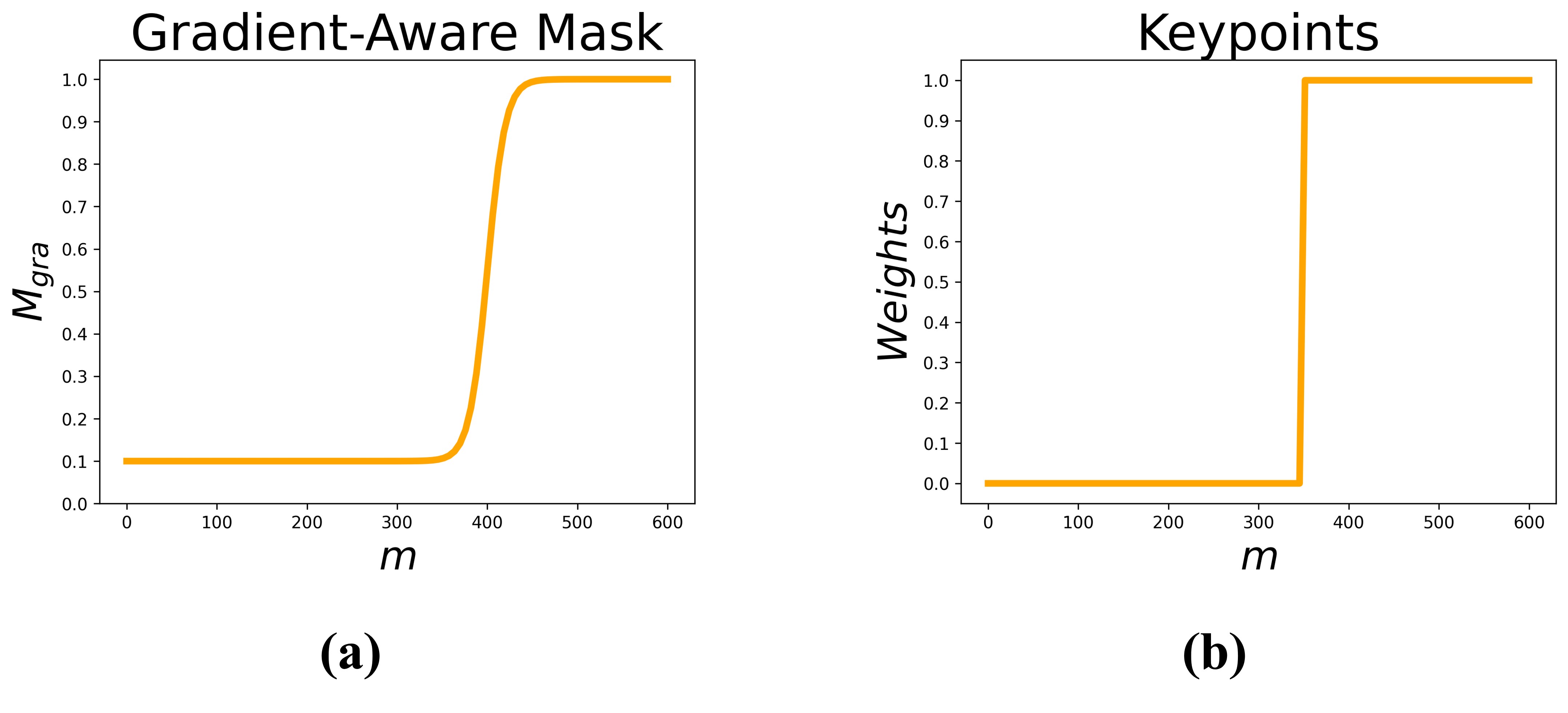}
\caption{Different weights assignment between our gradient-aware mask $\emph{\textbf{M}}_{gra}$ and keypoints-only photometric loss. (a) $\emph{\textbf{M}}_{gra}$ allocates different weights to various pixels based on their gradient magnitudes $\emph{\textbf{m}}$, providing more robust supervision for both textureless regions and key areas. (b) P\textsuperscript{2}Net~\cite{yu2020p} and StructDepth~\cite{li2021structdepth} allocate weight of 1 to keypoints and 0 to non-keypoints. }
\label{fig:mask function}
\end{figure}

\begin{figure}
\centering
\includegraphics[width=3in]{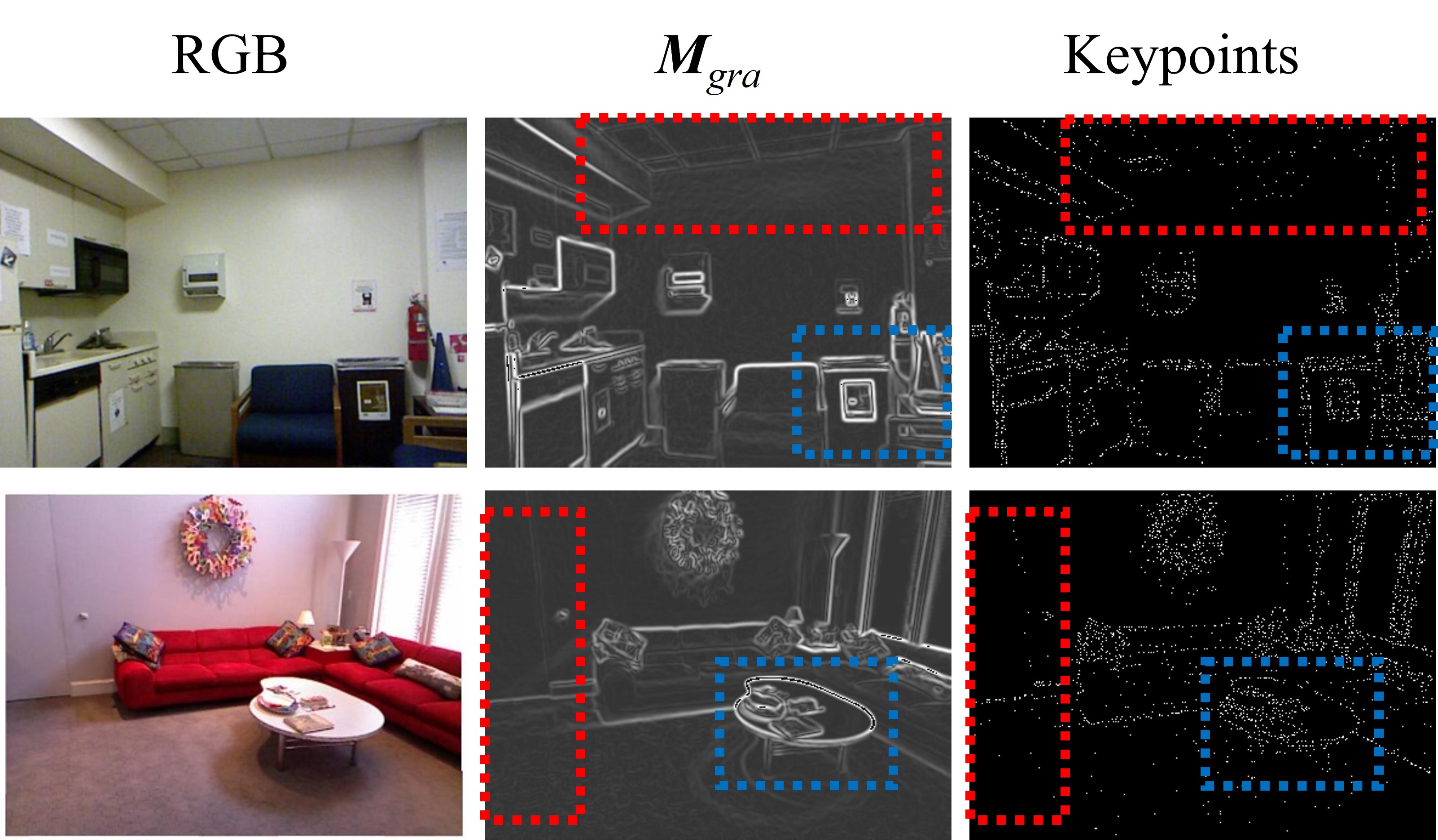}
\caption{Comparisons between our gradient-aware mask $\emph{\textbf{M}}_{gra}$ and keypoints detected by DSO~\cite{engel2017direct}. The gray level of each pixel corresponds to its weight, with closer to white indicating higher weights. Our $\emph{\textbf{M}}_{gra}$ provides adaptive supervision at both textureless (indicated by red boxes) and texture-rich regions (indicated by blue boxes). Weights of textureless regions are completely excluded in P\textsuperscript{2}Net~\cite{yu2020p} and StructDepth~\cite{li2021structdepth}.}
\label{fig:mask image}
\end{figure}

As shown in Fig.~\ref{fig:mask function}, $\emph{\textbf{M}}_{gra}$ adaptively and robustly allocates different weights to various pixels, contingent on their gradient magnitudes $\emph{\textbf{m}}$. Specifically, pixels with larger gradient magnitude, often indicating keypoints marked by significant RGB shift, receive higher weights during photometric loss calculation. Conversely, pixels with smaller gradient magnitudes, indicating textureless regions, are assigned lower weights, rather than being entirely ignored in P\textsuperscript{2}Net~\cite{yu2020p} and StructDepth~\cite{li2021structdepth}. Gradient magnitudes $\emph{\textbf{m}}$ are calculated using a 3x3 Sobel operator. Fig.~\ref{fig:mask image} demonstrates the difference between our gradient-aware mask $\emph{\textbf{M}}_{gra}$ and keypoints detected by DSO~\cite{engel2017direct}. The gray level of each pixel corresponds to its weight, with closer to white indicating higher weights. As shown in red boxes, in contrast to simply neglecting textureless regions, $\emph{\textbf{M}}_{gra}$ provides supervision for walls, ceilings, and floors.

Remarkably, $\emph{\textbf{M}}_{gra}$ not only elevates our baseline's performance but also enhances the depth estimation accuracy of other indoor self-supervised methods~\cite{sun2022sc} when combined with them. Details are presented in Sec.~\ref{sec:generalization ability}.

\subsection{Semantic Constraints}
\label{sec:semantic constraints}
Different from some established methods~\cite{klingner2020self},~\cite{tosi2020distilled} for outdoor scenes which utilize semantic and instance segmentation to detect dynamic objects, our approach pioneers the integration of semantic segmentation with depth estimation in an indoor self-supervised framework. Specifically, this is implemented through a shared encoder between DepthNet and SegNet, augmented by a cross-entropy segmentation loss $L_{seg}$ in the final loss function.

We employ the Light-weight RefineNet~\cite{nekrasov2018light}, a lightweight and pretrained semantic segmentation model, as our ProxySegNet. To follow the principles of self-supervised training, neither depth nor semantic segmentation ground-truth information is required during the training process. The ProxySegNet produces proxy semantic labels $\emph{\textbf{S}}_t^{proxy}$, guiding SegNet's predicted semantic labels $\emph{\textbf{S}}_t$ and refining depth estimation by capturing both low-level and high-level features through a shared encoder.


The cross-entropy segmentation loss $L_{seg}$ can be described as:
\begin{equation}
    L_{seg} = -\sum_c \sum_{p_t} \left( \emph{\textbf{S}}_t(p_t) \log(\emph{\textbf{S}}_t^{proxy}(c, p_t)) \right)
    \label{equation:segmentation loss}
\end{equation}
where $c$ denotes class labels.

The shared encoder empowers the model with holistic scene comprehension by capturing shared features. The decoder of SegNet has the same architecture as DepthNet’s decoder except for the last Softmax layer. Their weights are learned independently. To focus on depth training, $L_{seg}$ is reduced to 0 at the middle stage of training.

\subsection{Final Training Loss}

We combine the $L_{gra}$ and $L_{seg}$ with loss proposed in the baseline:
\begin{equation}
    L_{total} = L_{gra} + \lambda_1 L_{seg} + \lambda_2 L_{smooth} + \lambda_3 L_{norm} + \lambda_4 L_{planar}
    \label{equation:total loss}
\end{equation}
where $L_{smooth}$, $L_{norm}$ and $L_{planar}$ are the smoothness loss, the Manhattan normal loss, and the co-planar loss proposed by StructDepth~\cite{li2021structdepth}. $\lambda_1$, $\lambda_2$, $\lambda_3$, and $\lambda_4$ are the hyperparameter weights for corresponding losses.

\section{EXPERIMENTS}

In this section, we test the performance of GAM-Depth on NYUv2~\cite{silberman2012indoor} and its generalization ability on ScanNet~\cite{dai2017scannet} and InteriorNet~\cite{li2018interiornet}. We also conduct ablation studies to validate the effectiveness of each individual component.

\begin{table*}[h]
\renewcommand{\arraystretch}{1.3}
\caption{Qualitative Result on NYUv2~\cite{silberman2012indoor}. The best results in each category are in \textbf{bold}. $\downarrow$ indicates the lower the better while$\uparrow$ indicates the higher the better. \ding{51} denotes supervised learning, $\Delta$ denotes additional supervision from pretrained depth estimation model, and \ding{55} denotes self-supervised learning}
\label{tab: qualitative results on nyuv2}
\begin{center}
\begin{tabular}{l|c|c c|c c c}
\hline
Methods & Supervision & AbsRel $\downarrow$ & RMS $\downarrow$ & $\delta_1$ $\uparrow$ & $\delta_2$ $\uparrow$ & $\delta_3$ $\uparrow$ \\
\hline
Eigen et al.~\cite{eigen2014depth} & \ding{51} & 0.158 & 0.641 & 0.769 & 0.950 & 0.988\\ 
Laina et al.~\cite{laina2016deeper} & \ding{51} & 0.127 & 0.573 & 0.811 & 0.953 & 0.988\\ 
DORN~\cite{fu2018deep} & \ding{51} & 0.115 & 0.509 & 0.828 & 0.965 & 0.992\\
BTS~\cite{lee2019big} & \ding{51} & 0.110 & 0.392 & 0.885 & 0.987 & 0.994\\
P3Depth~\cite{patil2022p3depth} & \ding{51} & 0.104 & 0.356 & 0.898 & 0.981 & 0.996\\ 
NewCRFs~\cite{yuan2022neural} & \ding{51} & \textbf{0.095} & \textbf{0.334} & \textbf{0.922} & \textbf{0.992} & \textbf{0.998}\\
\hline
DistDepth~\cite{wu2022toward} & $\Delta$ & 0.130 & 0.517 & 0.832 & \textbf{0.963} & 0.990\\ 
SC-Depthv3~\cite{sun2022sc} & $\Delta$ & \textbf{0.123} & \textbf{0.486} & \textbf{0.848} & \textbf{0.963} & \textbf{0.991}\\
\hline
Monodepth2~\cite{godard2019digging} & \ding{55} & 0.161 & 0.600 & 0.771 & 0.948 & 0.987\\ 
P\textsuperscript{2}Net~\cite{yu2020p} & \ding{55} & 0.150 & 0.561 & 0.796 & 0.948 & 0.986\\ 
StructDepth~\cite{li2021structdepth} & \ding{55} & 0.142 & 0.540 & 0.813 & 0.954 & 0.988\\ 
Bian et al.~\cite{bian2021auto} & \ding{55} & 0.138 & 0.532 & 0.820 & 0.956 & 0.989\\ 
MonoIndoor~\cite{ji2021monoindoor} & \ding{55} & 0.134 & 0.526 & 0.823 & 0.958 & 0.989\\ 
HI-Net~\cite{wu2022hi} & \ding{55} & \textbf{0.130} & 0.522 & 0.828 & 0.959 & 0.989\\ 
Ours(GAM-Depth) & \ding{55} & 0.131 & \textbf{0.507} & \textbf{0.836} & \textbf{0.960} & \textbf{0.990}\\
\hline
\end{tabular}
\end{center}
\end{table*}

\begin{figure*}
\centering
\includegraphics[width=6in]{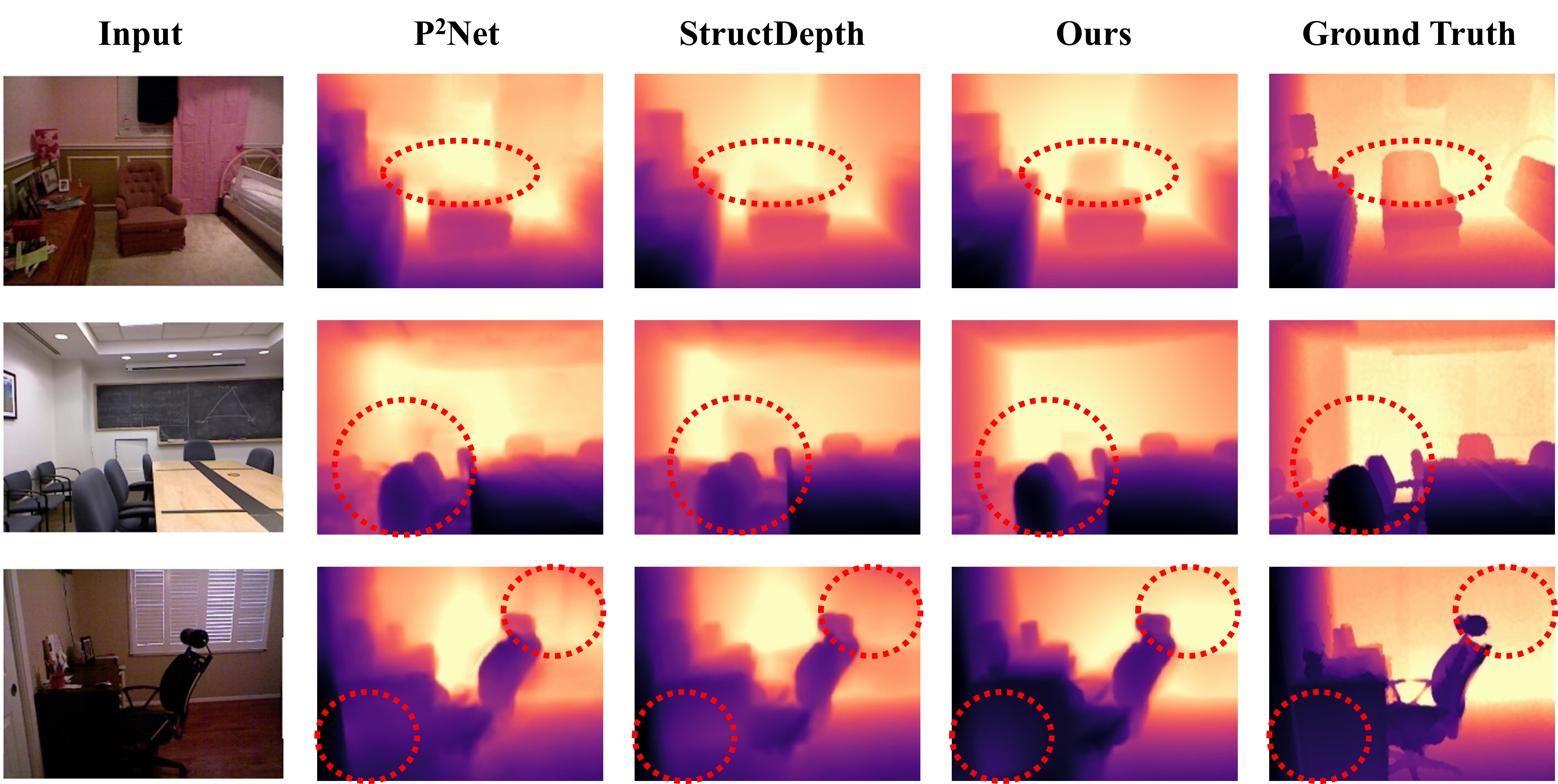}
\caption{Qualitative results on NYUv2~\cite{silberman2012indoor}. RGB images, P\textsuperscript{2}Net~\cite{yu2020p}, StructDepth~\cite{li2021structdepth}, our predictions, and ground truth depth maps are presented for comparison. GAM-Depth obtains more accurate results at object boundaries and textureless regions, as indicated by the red circles.}
\label{fig:qualitative results}
\end{figure*}

\subsection{Datasets}

Following P\textsuperscript{2}Net~\cite{yu2020p} and StructDepth~\cite{li2021structdepth}, we evaluate our model using three publicly available indoor scene datasets: NYUv2, ScanNet, and InteriorNet. Our training is based on NYUv2.

\subsubsection{NYUv2}
The NYUv2 dataset comprises 582 indoor scenes captured by a Microsoft Kinect camera. Following the split of~\cite{eigen2014depth} and~\cite{li2021structdepth}, 283 scenes (approximately 230K images) are used for training. The training set is sampled at intervals of 10 frames, using $\pm$10 and $\pm$20 frames as reference frames. After sampling, the total number of training samples is reduced to 21465. Following the official split, the validation set contains 654 dense labeled RGB-depth pairs.

\subsubsection{ScanNet}
ScanNet comprises approximately 2.5 million RGBD videos captured across 1513 scenes using a depth camera attached to an iPad.

\subsubsection{InteriorNet}
InteriorNet is a synthetic dataset containing millions of video sequences representing various interior design layouts, furniture, and object models.

\begin{table}[h]
\renewcommand{\arraystretch}{1.3}
\caption{ScanNet~\cite{dai2017scannet} results with the trained model on NYUv2~\cite{silberman2012indoor}}
\label{tab:scannet}
\begin{center}
\begin{tabular}{l|c c|c c c}
\hline
Methods & AbsRel $\downarrow$ & RMS $\downarrow$ & $\delta_1$ $\uparrow$ & $\delta_2$ $\uparrow$ & $\delta_3$ $\uparrow$ \\
\hline
Monodepth2~\cite{godard2019digging} & 0.191 & 0.451 & 0.693 & 0.926 & 0.983\\
P\textsuperscript{2}Net~\cite{yu2020p} & 0.175 & 0.420 & 0.740 & 0.932 & 0.982\\ 
StructDepth~\cite{li2021structdepth} & 0.165 & 0.400 & 0.754 & \textbf{0.939} & \textbf{0.985}\\
GAM-Depth & \textbf{0.159} & \textbf{0.390} & \textbf{0.770} & \textbf{0.939} & \textbf{0.985}\\ 
\hline
\end{tabular}
\end{center}
\end{table}

\begin{table}[h]
\renewcommand{\arraystretch}{1.3}
\caption{InteriorNet~\cite{li2018interiornet} results with the trained model on NYUv2~\cite{silberman2012indoor}}
\label{tab:interiornet}
\begin{center}
\begin{tabular}{l|c c|c c c}
\hline
Methods & AbsRel $\downarrow$ & RMS $\downarrow$ & $\delta_1$ $\uparrow$ & $\delta_2$ $\uparrow$ & $\delta_3$ $\uparrow$ \\
\hline
Monodepth2~\cite{godard2019digging} & 0.368 & 0.817 & 0.586 & 0.815 & 0.898\\
P\textsuperscript{2}Net~\cite{yu2020p} & 0.346 & 0.737 & 0.642 & 0.833 & 0.902\\ 
StructDepth~\cite{li2021structdepth} & 0.330 & 0.715 & 0.660 & 0.840 & 0.905\\
GAM-Depth & \textbf{0.313} & \textbf{0.693} & \textbf{0.676} & \textbf{0.846} & \textbf{0.907}\\ 
\hline
\end{tabular}
\end{center}
\end{table}

\subsection{Implementation Details}

The model is implemented using PyTorch~\cite{paszke2017automatic}. To allow a fair comparison with previous models, GAM-Depth adopts the implementation of StructDepth~\cite{li2021structdepth}, training the network for 50 epochs based on P\textsuperscript{2}Net~\cite{yu2020p}. We adopt Adam~\cite{kingma2014adam} as the optimizer. Due to limited computing resources, the batch size is decreased from 32 to 16, which impacts the results negatively. For fair comparisons with existing indoor self-supervised methods, the backbone model adopts ResNet18~\cite{he2016deep} and pose networks follow MonoIndoor~\cite{ji2021monoindoor}. Following~\cite{godard2019digging},~\cite{yu2020p} and~\cite{li2021structdepth}, we normalize the depth to 10m and use the median scaling strategy to avoid the scale ambiguity of self-supervised monocular depth estimation. Hyperparameter $\alpha$ for photometric loss calculation is set to 0.85 as~\cite{godard2017unsupervised},~\cite{godard2019digging}. Our gradient-aware photometric loss $L_{gra}$ weighting factors $\beta$ is set to 0.1 to balance the weights between textureless and texture-rich regions while $\gamma_1$ and $\gamma_2$ are set to 0.1 and 40, respectively. Hyperparameters $\lambda_1$, $\lambda_2$, $\lambda_3$, and $\lambda_4$ for the final training loss are set to 0.001, 0.1, 0.05, 0.1, respectively.

\subsection{Depth Estimation Results}

\subsubsection{Quantitative Results}
According to the quantitative results on NYUv2~\cite{silberman2012indoor} shown in Tab.~\ref{tab: qualitative results on nyuv2}, GAM-Depth achieves the best result and outperforms existing self-supervised indoor depth estimation methods by a large margin, especially on $\delta_1$ accuracy. Even if GAM-Depth is trained in the self-supervised mode, it outperforms some supervised and pretrained model-supervised methods. Tab.~\ref{tab:scannet} and Tab.~\ref{tab:interiornet} show the depth estimation results on ScanNet~\cite{dai2017scannet} and InteriorNet~\cite{li2018interiornet} with the model trained on NYUv2. Although they have not been used for training, the results show that GAM-Depth has better generalization ability than existing methods.

\begin{table}[h]
\scriptsize
\renewcommand{\arraystretch}{1.3}
\caption{Ablation studies on our gradient-aware mask $\emph{\textbf{M}}_{gra}$ and semantic constraints $L_{seg}$. Ours(key) and ours(avg) represent replacing $\emph{\textbf{M}}_{gra}$ with only considering keypoints' photometric losses or averaging all pixels' losses}
\label{tab:ablation of contribution}
\begin{center}
\begin{tabular}{l|c|c|c c|c c c}
\hline
Methods & $\emph{\textbf{M}}_{gra}$ & $L_{seg}$ & AbsRel $\downarrow$ & RMS $\downarrow$ & $\delta_1$ $\uparrow$ & $\delta_2$ $\uparrow$ & $\delta_3$ $\uparrow$ \\
\hline
Ours & \ding{55} & \ding{55} & 0.136 & 0.526 & 0.824 & 0.958 & 0.989\\ 
Ours & \ding{51} & \ding{55} & 0.134 & 0.520 & 0.828 & 0.959 & 0.989\\ 
Ours(key) & \ding{55} & \ding{51} & 0.133 & 0.516 & 0.830 & 0.959 & 0.989\\ 
Ours(avg) & \ding{55} & \ding{51} & 0.132 & 0.510 & 0.834 & \textbf{0.960} & \textbf{0.990}\\ 
\hline
Ours(full) & \ding{51} & \ding{51} & \textbf{0.131} & \textbf{0.507} & \textbf{0.836} & \textbf{0.960} & \textbf{0.990}\\ 
\hline
\end{tabular}
\end{center}
\end{table}

\begin{table}[h]
\scriptsize
\renewcommand{\arraystretch}{1.3}
\caption{Ablation studies on the generalization ability of our gradient-aware mask $\emph{\textbf{M}}_{gra}$}
\label{tab:ablation of mask}
\begin{center}
\begin{tabular}{l|c c|c c c}
\hline
Methods & AbsRel $\downarrow$ & RMS $\downarrow$ & $\delta_1$ $\uparrow$ & $\delta_2$ $\uparrow$ & $\delta_3$ $\uparrow$ \\
\hline
StructDepth~\cite{li2021structdepth} & \textbf{0.142} & \textbf{0.540} & \textbf{0.813} & \textbf{0.954} & \textbf{0.988}\\ 
StructDepth~\cite{li2021structdepth} + $\emph{\textbf{M}}_{gra}$ & 0.150 & 0.567 & 0.796 & 0.949 & 0.986\\ 
\hline
Ours(baseline) & 0.136 & 0.526 & 0.824 & 0.958 & \textbf{0.989}\\ 
Ours(baseline) + $\emph{\textbf{M}}_{gra}$ & \textbf{0.134} & \textbf{0.520} & \textbf{0.828} & \textbf{0.959} & \textbf{0.989}\\
\hline
SC-DepthV3~\cite{sun2022sc} & 0.123 & 0.486 & 0.848 & \textbf{0.963} & \textbf{0.991}\\ 
SC-DepthV3~\cite{sun2022sc} + $\emph{\textbf{M}}_{gra}$ & \textbf{0.120} & \textbf{0.471} & \textbf{0.854} & \textbf{0.963} & \textbf{0.991}\\ 
\hline
\end{tabular}
\end{center}
\end{table}

\subsubsection{Qualitative Results}
Fig.~\ref{fig:qualitative results} showcases the qualitative results of NYUv2. The depth predictions of GAM-Depth, StructDepth~\cite{li2021structdepth}, and P\textsuperscript{2}Net~\cite{yu2020p} are processed with the same median scaling method. Results are better viewed by coloring. The red circles show that GAM-Depth produces accurate object boundaries and continuous depth in textureless regions.

\subsection{Ablation Studies}

\subsubsection{Effectiveness of $\textbf{M}_{gra}$ and $L_{seg}$}

Tab.~\ref{tab:ablation of contribution} shows the improvements of our gradient-aware mask $\emph{\textbf{M}}_{gra}$ and semantic constraints $L_{seg}$. Excluding either $\emph{\textbf{M}}_{gra}$ or semantic constraints $L_{seg}$ shows degradation. Our baseline achieves better performance compared to~\cite{li2021structdepth} through residual pose estimations~\cite{ji2021monoindoor}. Replacing our gradient-aware mask $\emph{\textbf{M}}_{gra}$ with either only considering keypoints or averaging all pixels' losses shows deficiencies.

\subsubsection{Generalization ability of $\textbf{M}_{gra}$}
\label{sec:generalization ability}
In this part of the experiments, we appply our $\emph{\textbf{M}}_{gra}$ to a couple of photometric loss-based self-supervised depth estimation methods, including StructDepth~\cite{li2021structdepth}, and SC-DepthV3~\cite{sun2022sc}. As shown in  Tab.~\ref{tab:ablation of mask}, the use of $\emph{\textbf{M}}_{gra}$ elevates the performance of our baseline and SC-DepthV3, but degrades the performance of StructDepth. The degradation in StructDepth may be due to its limited supervision~\cite{li2021structdepth} whereas the improvement in SC-DepthV3~\cite{sun2022sc} is because of depth guidance from an off-the-shelf depth estimation model, which serves as a strong supervision signal. 

\section{CONCLUSIONS}

In this paper, we have developed GAM-Depth for self-supervised indoor depth estimation, built upon our proposed gradient-aware mask and incorporation of semantic constraints. The gradient-aware mask provides robust supervision for both textureless and texture-rich regions by adaptive weight allocation in the photometric loss calculation. The semantic constraints enable a co-optimization training pipeline that integrates low-level details and high-level scene comprehension. Results have proved that our model can produce smooth depth in textureless regions and sharp depth at boundaries.

\newpage

\addtolength{\textheight}{-12cm}   











\bibliographystyle{IEEEtran}
\bibliography{bio.bib}

\end{document}